%% file: acl_latex.tex
\pdfoutput=1
\documentclass[11pt]{article}

\usepackage[final]{acl}

\usepackage{times}
\usepackage{latexsym}

\usepackage[T1]{fontenc}
\usepackage[utf8]{inputenc}

\usepackage{microtype}

\usepackage{inconsolata}

\usepackage{graphicx}
\usepackage{amsmath}
\usepackage{amssymb}
\usepackage{algorithm}
\usepackage{algorithmic}
\usepackage{multirow}
\usepackage{mdframed}
\usepackage{adjustbox}
\usepackage{booktabs}
\usepackage{tcolorbox}
\usepackage{xcolor}
\usepackage{float}
\usepackage{tabularray}
\usepackage{arydshln}
\usepackage{xspace}
\usepackage{pifont}% http://ctan.org/pkg/pifont
\newcommand{\cmark}{\ding{51}}%
\newcommand{\xmark}{\ding{55}}%
\usepackage{colortbl}
\newmdenv[backgroundcolor=gray!10, linewidth=0pt, innerleftmargin=10pt, innerrightmargin=10pt, innertopmargin=10pt, innerbottommargin=10pt]{highlight}

\definecolor{mypink1}{rgb}{0.858, 0.188, 0.478}
\definecolor{codegreen}{rgb}{0,0.6,0}
\definecolor{codegray}{rgb}{0.5,0.5,0.5}
\definecolor{codepurple}{rgb}{0.58,0,0.82}
\definecolor{backcolour}{rgb}{0.95,0.95,0.92}

\usepackage[normalem]{ulem}
\useunder{\uline}{\ul}{}

\usepackage{listings}
\lstdefinestyle{mystyle}{
    backgroundcolor=\color{backcolour}, 
    commentstyle=\color{codegreen},
    keywordstyle=\color{magenta},
    numberstyle=\tiny\color{codegray},
    stringstyle=\color{codepurple},
    basicstyle=\ttfamily\footnotesize,
    breakatwhitespace=false,         
    breaklines=true,                 
    captionpos=b,                    
    keepspaces=true,                 
    numbers=left,                    
    numbersep=5pt,                  
    showspaces=false,                
    showstringspaces=false,
    showtabs=false,                  
    tabsize=2
}

\lstdefinelanguage{JSON}{
    keywords={true,false,null},
    ndkeywords={if,else,switch,case,break,default,for,while,do,return,void,const,int,float,double,char,string,long,short,try,catch,throw,throws,public,private,protected,static,abstract,interface,new,class,implements,extends,assert,true,false,null},
    morestring=[b]',
    morestring=[b]",
    morecomment=[l]//,
    morecomment=[s]{/*}{*/},
    morecomment=[s]{/**}{*/},
    morecomment=[s]{/*!}{*/},
}

\newcommand{\name}[0]{\textsc{ToolFlow}\xspace}

\title{\textsc{ToolFlow}: Boosting LLM Tool-Calling\\Through Natural and Coherent Dialogue Synthesis}
\author{
  Zezhong Wang$^{1}$\thanks{Work done during internship at Huawei Noah’s Ark Lab.}, Xingshan Zeng$^{2}$\thanks{Corresponding author}, Weiwen Liu$^{2}$, Liangyou Li$^{2}$, \\ 
  \bf Yasheng Wang$^{2}$, Lifeng Shang$^{2}$, Xin Jiang$^{2}$, Qun Liu$^{2}$, Kam-Fai Wong$^{1}$ \\
  $^1$The Chinese University of Hong Kong,  $^2$Huawei Noah’s Ark Lab\\
  \texttt{\{zzwang,kfwong\}@se.cuhk.edu.hk}\\
  \texttt{\{zeng.xingshan,liuweiwen8,liliangyou\}@huawei.com}\\
  \texttt{\{wangyasheng,Shang.Lifeng,Jiang.Xin,qun.liu\}@huawei.com}
  }

\begin{document}
\maketitle
\begin{abstract}
Supervised fine-tuning (SFT) is a common method to enhance the tool calling capabilities of Large Language Models (LLMs), with the training data often being synthesized. The current data synthesis process generally involves sampling a set of tools, formulating a requirement based on these tools, and generating the call statements. However, tools sampled randomly lack relevance, making them difficult to combine and thus reducing the diversity of the data. Additionally, current work overlooks the coherence between turns of dialogues, leading to a gap between the synthesized data and real-world scenarios. 
To address these issues, we propose a Graph-based Sampling strategy to sample more relevant tool combinations, and a Planned-generation strategy to create plans that guide the synthesis of coherent dialogues. 
We integrate these two strategies and enable multiple agents to synthesize the dialogue data interactively, resulting in our tool-calling data synthesis pipeline \name. Data quality assessments demonstrate improvements in the naturalness and coherence of our synthesized dialogues.
Finally, we apply SFT on LLaMA-3.1-8B using 8,000 synthetic dialogues generated with \name. Results show that the model achieves tool-calling performance comparable to or even surpassing GPT-4, while maintaining strong general capabilities.
\end{abstract}

\input{chapters/introduction}

\input{chapters/related_work}

\input{chapters/methodology}
\input{chapters/experiment}
\input{chapters/analysis}

\input{chapters/overlap_analysis}
\input{chapters/conclusion}

\input{chapters/limitation}

\input{chapters/ethic_statement}
\input{chapters/acknowledgement}

\bibliography{acl_latex}

\input{chapters/appendix}

\end{document}

%% file: chapters/introduction.tex
\section{Introduction}
Enabling Large Language Models (LLMs) to perform tool calling significantly enhances their capabilities and practical applications. This requires the models to possess strong understanding, reasoning, and instruction-following abilities.
Customized fine-tuning is a widely used method to improve the tool-calling capabilities of LLMs~\cite{abdelaziz2024granitefunctioncallingmodelintroducing, patil2023gorillalargelanguagemodel, schick2023toolformerlanguagemodelsteach, qin2023toolllmfacilitatinglargelanguage}. However, access to fine-tuning data can be limited. One viable solution is to utilize LLMs for data synthesis~\cite{basu-etal-2024-api, wang2023selfinstructaligninglanguagemodels, xu2023wizardlmempoweringlargelanguage, yu2024metamathbootstrapmathematicalquestions}.

A typical tool-calling data synthesis process involves three steps: (1) selecting candidate tool(s), (2) generating requirements based on those tools, and (3) creating the call statements~\cite{tang2023toolalpacageneralizedtoollearning, liu2024apigenautomatedpipelinegenerating}. However, the data synthesized through this method often lacks realism and naturalness. Randomly sampled tools frequently fail to interconnect, making it difficult to combine them for complex tasks. Consequently, the requirements for subsequent synthesis tend to be simplistic, which reduces the diversity and complexity of the data. Furthermore, much of the existing research focuses solely on generating single-turn tool-calling instructions, neglecting the coherence between dialogue turns~\cite{qin2023toolllmfacilitatinglargelanguage,yang2023gpt4toolsteachinglargelanguage}. In real-world interactions, LLMs typically engage with users through dialogues rather than single-round Q\&A sessions. This creates a gap between Q\&A-type training data and its practical application, ultimately diminishing the naturalness of the synthesized data.

To address these two challenges, we propose \name, a tool-calling data synthesis pipeline that employs a graph-based sampling algorithm to improve the correlation among the selected tools and a planned-generation strategy to enhance the naturalness and coherence of the synthesized tool call dialogues.

Specifically, we consider tools with similar parameters or return values to be related. For instance, both "\textit{book\_flight}" and "\textit{get\_weather}" require parameters related to location. In practical scenarios, these two tools are indeed interconnected, as they often occur together in travel contexts. Based on this assumption, we construct a tool graph that represents the similarity between parameters and return values of the tools. Each node in the graph represents a tool, while the edges indicate the relevance between pairs of tools. When sampling tools, we randomly select a subgraph from this tool graph, ensuring that the sampled tools are more likely to interact effectively, thereby facilitating the generation of complex requirements.

On the other hand, before synthesizing dialogues, we first have the LLM create a plan based on the selected subset of tools. This plan outlines the requests that users need to make in each turn of the dialogue. While constructing the plan, the model focuses on establishing the dialogue framework without worrying about phrasing and details. This approach allows the model to concentrate on the logical relationships and interactions between requirements, resulting in more coherent demands.
Additionally, we enable the LLM to incorporate non-tool-call requests into the plan. This not only enhances the diversity of the conversation content but also facilitates seamless transitions between topics, naturally leading to new requirments.

We generate dialogues using three agents: \textit{User}, \textit{Assistant}, and \textit{Tool}. Based on the selected tool subset and the established plan, these agents interact to complete the dialogue. By iterating through the "sampling-planning-generation" process, we synthesized a total of 8,000 dialogues.
To evaluate the effectiveness of our proposed method, we conduct a comprehensive ablation study on the graph-based sampling and planning strategy by generating dialogues of the same size selectively without these modules. We perform a thorough evaluation of the data quality, which demonstrates that \name \space can effectively enhance the naturalness, coherence, and diversity of the generated dialogues. Finally, we apply supervised fine-tuning to LLaMA-3.1-8B-Instruct~\cite{dubey2024llama3herdmodels} using the synthesized data and validate improvements in the model's tool-calling capabilities while preserving its general abilities, with \name.

We summarize our contributions into the following three key points: \vspace{-0.25cm}
\begin{itemize}
\item We propose a Graph-based Sampling strategy to select related tools, aiming to enhance the diversity and complexity of synthetic tool calling requirements.
\item We introduce a Planned-Generation strategy to improve the naturalness and coherence of synthetic dialogues.
\item We integrate these two strategies and propose \name, a tool calling dialogue data synthesis pipeline, with extensive experiments and analyses showing its effectiveness.
\end{itemize}

%% file: chapters/related_work.tex
\section{Related Works}
Integrating external tools with large language models (LLMs) significantly broadens their functional scope, allowing for more specialized, precise, and reliable solutions to complex problems~\cite{qin2023toolllmfacilitatinglargelanguage}. There are generally two main strategies for embedding tool-use capabilities into LLMs: prompt-based methods and tool-augmented SFT. Prompt-based methods enable LLMs to utilize tools by providing descriptions and examples of the tools in the prompt, without any incremental training~\cite{ruan2023tptulargelanguagemodelbased,hsieh2023tooldocumentationenableszeroshot, mialon2023augmentedlanguagemodelssurvey}. 
ReAct~\cite{yao2023reactsynergizingreasoningacting} is one of the notable methods within this category. It allows LLMs to switch between reasoning and executing actions to tackle challenging tasks. However, the effectiveness of these approaches can be limited by the inherent capabilities of the model.
On the other hand, tool-augmented tuning is attracting increasing interest due to its direct enhancement of LLM’s tool usage capabilities~\cite{abdelaziz2024granitefunctioncallingmodelintroducing, patil2023gorillalargelanguagemodel, schick2023toolformerlanguagemodelsteach, qin2023toolllmfacilitatinglargelanguage}. As the limited availability of tool calling datasets, ~\citet{basu-etal-2024-api} adapted data from various other domains for application in tool calling studies. Others~\cite{liu2024apigenautomatedpipelinegenerating, tang2023toolalpacageneralizedtoollearning, qin2023toolllmfacilitatinglargelanguage} mainly synthesized single-turn instructions that involve basic tool calling requirements. However, LLMs typically interact with users through dialogue rather than single-turn Q\&A. This mismatch means the data is unnatural, creating a gap with real-world scenarios. Our \name focuses on enhancing the coherence and naturalness of dialogues in data synthesis, making it more aligned with actual applications.

%% file: chapters/methodology.tex
\section{Methodology}

\begin{figure*}[t]
    \centering
    \includegraphics[width=\linewidth]{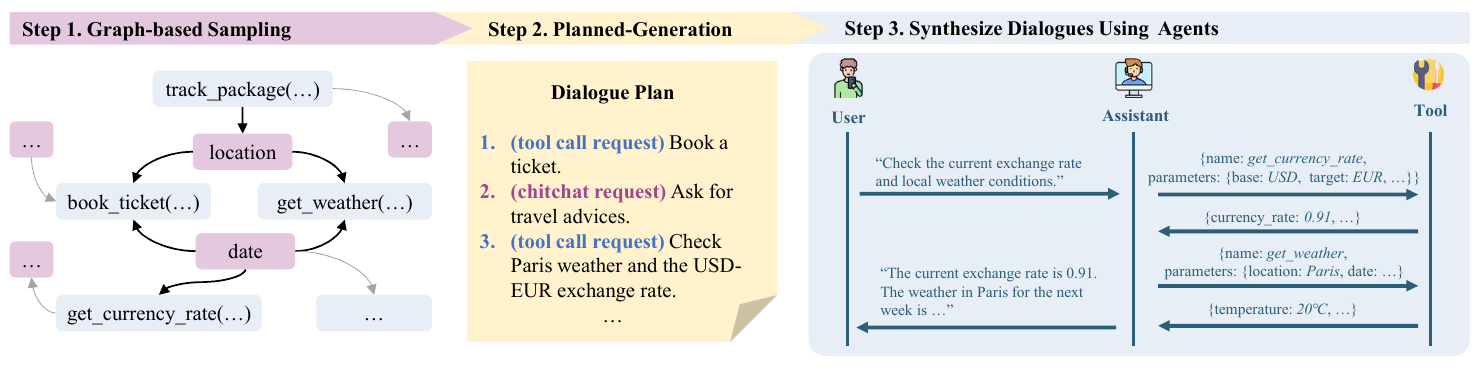}
    \caption{The pipeline of dialogue synthesis. The left side shows the Tool Graph with blue boxes representing tools and purple boxes representing parameters or return values. In the middle is the dialogue synthesis plan generated according to sampled tools. On the right is an example of data synthesis by the \textit{User}, \textit{Assistant}, and \textit{Tool} agents.}
    \label{fig:pipeline}
\end{figure*}

To generate realistic and coherent dialogues, we propose a three-step data synthesis process: (1) Selecting a tool subset using graph-based sampling; (2) Generating a dialogue plan based on the selected tool subset; (3) Synthesizing dialogues guided by the tool subset and dialogue plan.

\subsection{Graph-based Sampling for Tool Selection}
Tool calling data synthesis generally starts by selecting one or more tools from the available toolset. While much of the previous work overlooks the significance of tool selection~\cite{qin2023toolllmfacilitatinglargelanguage, patil2023gorillalargelanguagemodel}, relying solely on random sampling, the chosen tools play a crucial role in shaping the quality of the synthesized dialogues. In real-world scenarios, user requirements often necessitate the combined use of multiple tools to achieve a solution. To synthesize more complex user needs, it is vital to ensure that the selected tools can work together. To address this, we propose a Graph Sampling strategy to identify relevant and compatible tool combinations.

\subsubsection{Tool Graph Construction}

We first construct a graph, $G = (V, E)$, where node $v_i\in V$ represents the tool $i$, and the edge $e_{i,j}\in E$ represents whether tool $i$ and tool $j$ are related. The left side of Figure~\ref{fig:pipeline} shows an example of the tool graph.
We consider tools with similar parameters or return values to be related to each other:

\textbf{P-P Similarity:}
When two tools share similar parameters, there is a high probability that the tools are related. For instance, based on "\textit{location}" and "\textit{destination}", two semantically similar parameters, we can identify tools like "\textit{get\_weather}" and "\textit{book\_flight}", which are frequently used together in travel-related contexts.

\textbf{P-R Similarity:}
If the return value of one tool is similar to the input parameters of another, there is also a high likelihood that the two tools are related. For example, the \textit{"check\_calendar"} tool typically returns the location of events, while the \textit{"navigate"} tool requires a location as input. When a user requests to \textit{"navigate to the location of this afternoon's meeting,"} both tools would be called.

To derive similarity between parameters or return values, we first concatenate the name and description of a parameter or a return value using the template \textit{"\{Name\}: \{Description\}"}. For example, the parameter ``Date'' of one specific tool is represented as \textit{"Date: Departure date, format as dd/mm/yyyy."} 
Then, we encode these strings using \texttt{Sentence-BERT}~\cite{reimers2019sentencebertsentenceembeddingsusing} to obtain the corresponding embeddings.
We use $p^i_k$ and $r^i_l$ to denote the $k$-th parameters and $l$-th return values of the tool $v_i$, respectively. And we use  $\mathbf{p}^i_k$ and $\mathbf{r}^i_k$ to represent the embeddings of $p^i_k$ and $r^i_l$, respectively.
The similarity between $v_i$'s parameter $p^i_k$ and $v_j$'s parameter $p^j_l$ is defined as:
\begin{equation}
    \cos(\mathbf{p}^i_k, \mathbf{p}^j_l) = \frac{\mathbf{p}^i_{k} \cdot \mathbf{p}^j_{l}}{\|\mathbf{p}^i_{k}\| \|\mathbf{p}^j_{l}\|}
\end{equation}
Similarly, the similarity between $v_i$'s return value $r^i_k$ and $v_j$'s parameter $p^j_l$ is defined as $\cos(\mathbf{r}^i_k, \mathbf{p}^j_l)$.
If the similarity is greater than a predefined threshold $\tau$, we consider the two tools to be correlated. We set $\tau$ to be 0.82 according to our preliminary study. 
This means that when the similarity between any pair of parameters from two tools exceeds $\tau$, or when the similarity between a return value and an input parameter of two tools exceeds $\tau$,
we assign an edge between the two tools:
\begin{equation}
e_{i, j} = 
\begin{cases} 
    1, & \exists p^i_k\subseteq v_i, p^j_l\subseteq v_j : \cos(\mathbf{p}^i_k, \mathbf{p}^j_l) > \tau \\
    & \quad\quad\quad\quad\quad\text{or} \\
    & \exists r^i_k\subseteq v_i, p^j_l\subseteq v_j : \cos(\mathbf{r}^i_k, \mathbf{p}^j_l) > \tau \\
    0, & \text{otherwise}
\end{cases}
\end{equation}
where $i, j=1\cdots N$ and $i \neq j$. We use $\subseteq$ to represent a parameter or a return value is included in the tool.

\subsubsection{Graph-based Sampling}
With the constructed tool graph, we are able to sample a subset consisting of $n$ tools that might be correlated. Generally, we randomly select a node as the starting point on the graph, and then perform a random walk along the edges of the graph. We stop when the path length reaches $n$, and the nodes included in the path constitute the sampled subset of tools. Details are shown in Algorithm~\ref{alg:graph_sample}.

\input{tables/alg}

\subsection{Dialogue Plan Generation}
Coherent dialogues usually involve complex tool-calling scenarios, such as cases where the current tool calling relies on the return value of a previous one or where parameters for the current tool calling are already present in the dialogue history. Moreover, realistic dialogues between humans and AI assistants often don’t always require tool use; they frequently involve non-tool-related exchanges, such as chitchat, interspersed with tool-based tasks. To enhance an LLM's performance in such realistic scenarios, training examples that reflect this balance are essential.

To address the need and enhance the coherence of synthesized dialogues, we propose a Planned-Generation strategy, which indicates planning before generating. We have the LLM first formulates a set of user requirements based on the tool subset to create a dialogue plan. These requirements can involve tool call requests -- tasks that necessitate the use of these tools -- or non-tool interactions, such as chitchat. The middle part of Figure~\ref{fig:pipeline} shows an example of a plan. At this stage, the LLM focuses solely on the logic, coherence, and natural flow of the requirements, without delving into the phrasing of interactions or other nuances.
Compared to directly synthesized dialogues, those generated based on a dialogue plan show markedly improved coherence. We provide a detailed assessment of this coherence in the subsequent sections. Please refer to Table~\ref{tab:plan_prompt} for the plan synthetic prompt.

\subsection{Multi-Agent Dialogue Synthesis}

We set up three agents, \textit{user}, \textit{assistant}, and \textit{tool}, with LLMs to collaboratively synthesize dialogues. The right side of Figure~\ref{fig:pipeline} illustrates the synthesis process for one dialogue turn.

The \textit{user} agent is responsible for initiating requests based on the dialogue plan. It first checks whether the current request in the plan has been completed, which determines whether to continue with the current task or move on to the next one. This ensures that the dialogue stays aligned with the plan's flow and sequence.

The \textit{assistant} agent evaluates the user’s request to determine if a tool is required. In cases where no tool is needed, such as chitchat, the assistant agent responds directly. If a tool call is necessary, the assistant verifies whether all required parameters are present based on the tool documentation. If any parameters are missing, the assistant requests clarification from the user; otherwise, it generates the tool call statement.

The \textit{tool} agent simulates the return values of the requested tool based on the tool documentation and the assistant's call statement. 

The interaction among the three agents continues for each dialogue turn until all the requests in plan are addressed or the preset turn limit is reached.
Afterward, all dialogue turns are collected and a rule-based data filtering module is applied to remove low-quality data~\cite{liu2024apigenautomatedpipelinegenerating, liu2024toolacewinningpointsllm}. 
The filtering rules primarily check the format of tool call statements, as well as other issues such as incomplete dialogues or missing tool call turns.

\subsection{Implementation Details}
\label{sec:imp_details}

In this work, for tool selection, we directly utilize the tools from ToolBench~\cite{qin2023toolllmfacilitatinglargelanguage} (including over 16,000 RESTful APIs) as our available tools. To standardize tool descriptions, we follow the setting of OpenAI Function Calling\footnote{https://platform.openai.com/docs/guides/function-calling} and prompt an LLM (Llama-3.1-8B) to convert all these tools into JSON format. For cases where information is incomplete, such as missing parameter descriptions, we rely on the LLM to infer and fill in the missing details during the conversion process. A demonstration tool is presented in Figure~\ref{fig:tool_demo}.

We use GPT-4~\cite{openai2024gpt4} for all generative tasks, including dialogue plan generation and agent simulation, unless otherwise specified. Multiple versions of GPT-4 are randomly selected for each dialogue, potentially enhancing diversity.

\input{tables/statistics}

\section{Data Quality Assessment}
\subsection{Basic Data Information}
In this section, we evaluate the quality of the synthetic data. To assess the effectiveness of the Graph-based Sampling (referred to as Graph) and the Planned-Generation strategy (referred to as Plan), we synthesized three additional sets of comparative data under different conditions: removing Graph, removing Plan, and removing both Graph and Plan. Each dataset contains 8,000 dialogues. Table~\ref{tab:statistics} presents the total number of tokens, the number of tool calls, and the number of dialogue turns containing tool calls for these datasets.

As shown in Table~\ref{tab:statistics}, the total number of tokens generated by the different strategies is similar, at approximately 8 million. Dialogues synthesized using the Planned-Generation strategy include more non-tool interactions, resulting in a lower proportion of tool call requests. In contrast, the Graph-based Sampling strategy increases the number of tool calls. 
This can be attributed to the connections among tools, where relevant information for subsequent tool calls is generally contained in the dialogue history, thereby reducing the need for additional turns to ask for missing information.

\subsection{Quality Evaluation}
\label{sec:quality_evaluation}

To further assess the quality of the synthetic dialogues, we implemented both an automatic evaluation and a model-based evaluation. 

The automatic evaluation primarily assesses the coherence and diversity of the dialogues. 
Following~\citet{dziri-etal-2019-evaluating}, we assess the coherence of the dialogue as a Natural Language Inference (NLI) task. 
We treat two consecutive turns in the dialogue as the premise and hypothesis, respectively, and calculate the ratio of entailment relation (\textit{EnR}) as well as the semantic similarity (\textit{SS}) between them. A higher EnR or SS between turns indicates that the dialogue is more coherent.

Regarding diversity, we calculate the text's Shannon entropy (\textit{H}) based on the word frequency~\cite{6773024}. 
We also compute the Distinct-N Score~\cite{li-etal-2016-diversity} for the dataset, with $N=3$ (\textit{D-3}). Higher entropy or Distinct-N Score indicates that the dataset contains more information and has greater diversity.

In addition, we randomly sampled 200 dialogues in each dataset for the model-based evaluation. We used GPT-4~\cite{openai2024gpt4} to carefully evaluate each dialogue based on four dimensions: naturalness (\textit{NAT}), coherence (\textit{COH}), helpfulness (\textit{HELP}), and accuracy (\textit{ACC}). The prompt for GPT-4 evaluation is shown in Table~\ref{tab:eval_prompt}.

The evaluation results are shown in Table~\ref{tab:data_quality}. There are two key observations:
\begin{itemize}\vspace{-0.2cm}
    \item \textit{H} and \textit{D-3} Score demonstrate that Graph Sampling enhances the diversity of the data.\vspace{-0.2cm}
    \item Both evaluations (\textit{SS}, \textit{EnR}, and \textit{COH}) show that Planned-generation improves the coherence of the dialogue. \vspace{-0.2cm}
\end{itemize}

For more detailed settings, analysis and explanations, please refer to Appendix~\ref{sec:data_quality_assess_app}.

\input{tables/data_eval}

\subsection{Comparison with Natural Dialogue Dataset}
\label{sec:comp_with_ds}
To better understand how our synthetic dialogues compare with human-created ones, we conducted a comparative study with an established dataset. We chose the MultiWOZ dataset~\cite{budzianowski-etal-2018-multiwoz} as a natural dialogue dataset for comparison. MultiWOZ (Multi-Domain Wizard-of-Oz) is a well-known task-oriented dialogue dataset, and we believe comparisons with this dataset would be convincing. We repeated the GPT-4 evaluation experiment by first randomly sampling 200 dialogues from MultiWOZ. Then, using the same scoring prompt, we had GPT-4 evaluate these dialogues across the four dimensions. The results are shown in Table~\ref{tab:multi_woz_comp}.

MultiWOZ scores slightly higher on naturalness, coherence, and accuracy compared to the ToolFlow dataset, though the differences are minimal (average score differences between 0.1-0.2). Regarding helpfulness scores, ToolFlow outperformed MultiWOZ by 0.3 points. These results suggest that our synthetic dialogues in \textsc{ToolFlow} achieve comparable quality to human-created dialogues in MultiWOZ, with particularly strong performance in task-oriented aspects such as helpfulness. 
\input{tables/multi_woz_comp}

%% file: tables/alg.tex
\begin{algorithm}[t]
\caption{Graph-based Sampling}
\label{alg:graph_sample}
\small
\begin{algorithmic}[1]
\REQUIRE $G = (V, E)$ with $|V| = N$, integer $n \leq N$ (desired sample size)
\ENSURE Subset $\hat{V} \subseteq V$ with $|\hat{V}| = n$
\STATE {\textcolor{mypink1}{\textit{// Randomly choose a node from $V$ and add it to $\hat{V}$}}}
% \STATE $v_i \leftarrow \text{Uniform}(V)$ 
\STATE $\hat{V} \leftarrow \{\text{Uniform}(V)\}$ 
\WHILE{$|\hat{V}| < n$}
    % \IF{$\hat{V} = \emptyset$}
        % \STATE {\textit{\% Randomly choose a node from $V$}}
        % \STATE $v_i \leftarrow \text{Uniform}(V)$ 
        % \COMMENT{Choose a node uniformly at random from $V$}
        
    % \ELSE
    \STATE $v_i \leftarrow \text{last element of } \hat{V}$
    % \ENDIF
    \STATE {\textcolor{mypink1}{\textit{// Find all neighbors of $v_i$}}}
    \STATE $N(v_i) \leftarrow \{v_j \in V \mid e_{i, j} = 1, \forall e_{i, j} \in E\}$ 
    % \COMMENT{Find all neighbors of $v_i$}
    \STATE {\textcolor{mypink1}{\textit{// Randomly choose a neighbor of $v_i$}}}
    \STATE $v_j \leftarrow \text{Uniform}(N(v_i))$ 
    % \COMMENT{Choose a neighbor of $v$ uniformly at random}
    \IF{$v_j \notin \hat{V}$}
        \STATE {\textcolor{mypink1}{\textit{// Add $v_j$ to $\hat{V}$ if not already included}}}
        \STATE $\hat{V} \leftarrow \hat{V} \cup \{v_j\}$ 
        % \COMMENT{Add $u$ to $\hat{V}$ if not already present}
    \ENDIF
\ENDWHILE
\RETURN $\hat{V}$
\end{algorithmic}
\end{algorithm}

%% file: tables/statistics.tex
\begin{table}[t]
\centering\small
\begin{tabular}{llccc}
\hline
\rowcolor[HTML]{EFEFEF}\multicolumn{2}{c}{\textbf{Setting}} & \multicolumn{3}{c}{\textbf{Statistics}}                     \\ 
\textit{Graph}    & \textit{Plan}    & \textit{\# tokens} & \textit{\# call} & \textit{\# call turn} \\ \cmidrule(lr){1-2} \cmidrule(lr){3-5}
\cmark             & \cmark            & 8,054,298        & 21,069           & 17,112                \\
\cmark             & \xmark            & 8,145,545        & 25,158           & 21,504                \\
\xmark             & \cmark            & 7,956,087        & 18,117           & 15,931                \\
\xmark             & \xmark            & 8,069,304        & 23,301           & 20,804                \\ \hline
\end{tabular}
\caption{Basic information about the dialogue datasets synthesized by \name \space and its ablation settings.
\textit{\# token}, \textit{\# call}, and \textit{\# call turn} represent the number of tokens, tool calls, and turns containing tool calls in the dataset, respectively.}
\label{tab:statistics}
\end{table}

%% file: tables/data_eval.tex
\begin{table}[t]
\centering\small
\begin{tabular}{llcccc}
\hline
\rowcolor[HTML]{EFEFEF}\multicolumn{6}{c}{\textbf{Automatic Evaluation}}                                                                                    \\
\multicolumn{2}{c}{\textit{Setting}} & \multicolumn{2}{c}{\textit{Coherence}}              & \multicolumn{2}{c}{\textit{Diversity}}  \\ 
\cmidrule(lr){1-2} \cmidrule(lr){3-4} \cmidrule(lr){5-6}
\textit{Graph}    & \textit{Plan}    & \textit{SS}              & \textit{EnR}            & \textit{H}               & \textit{D-3} \\
\cmark             & \cmark            & 63.36                    & 47.3                     & 10.36                    & 0.4865       \\
\cmark             & \xmark            & 62.03                    & 32.1                     & 10.14                    & 0.4364       \\
\xmark             & \cmark            & 61.72                    & 48.1                     & 9.82                     & 0.3393       \\
\xmark             & \xmark            & 58.65                    & 35.4                     & 9.75                     & 0.3078       \\
\rowcolor[HTML]{EFEFEF}\multicolumn{6}{c}{\textbf{GPT-4 Evaluation}}                                                                                        \\
\textit{Graph}    & \textit{Plan}    & \textit{NAT}             & \textit{COH}             & \textit{HELP}            & \textit{ACC} \\
\cmark             & \cmark            & \multicolumn{1}{l}{3.72} & \multicolumn{1}{l}{3.91} & \multicolumn{1}{l}{4.71} & 4.92         \\
\cmark             & \xmark            & \multicolumn{1}{l}{3.00} & \multicolumn{1}{l}{3.72} & \multicolumn{1}{l}{4.51} & 4.84         \\
\xmark             & \cmark            & \multicolumn{1}{l}{3.51} & \multicolumn{1}{l}{3.88} & \multicolumn{1}{l}{4.39} & 4.87         \\
\xmark             & \xmark            & \multicolumn{1}{l}{2.93} & \multicolumn{1}{l}{3.66} & \multicolumn{1}{l}{4.18} & 4.90         \\ \hline
\end{tabular}
\caption{Results of automatic evaluation and GPT-4 evaluation on the data synthesized by \name \space and its ablation settings.}
\label{tab:data_quality}
\end{table}

%% file: tables/multi_woz_comp.tex
\begin{table}[t]
\centering
\begin{tabular}{lcccc}
\hline
\rowcolor[HTML]{EFEFEF}\textbf{Dataset} & \textbf{NAT} & \textbf{COH} & \textbf{HELP} & \textbf{ACC} \\ 
% \hline
\textsc{ToolFlow}         & 3.72         & 3.91         & 4.71          & 4.92         \\
MultiWOZ         & 3.98         & 4.03         & 4.41          & 4.95         \\ \hline
\end{tabular}
\caption{Evaluation Scores Comparison between \textsc{ToolFlow} and MultiWOZ.}
\label{tab:multi_woz_comp}
\end{table}

%% file: chapters/experiment.tex
\section{Experiments}
\label{sec:experiments}
\subsection{Settings}

% \input{tables/bfcl}
\input{tables/bfcl-v1v2}

\subsubsection{Datasets} We conducted experiments on the following three tool-calling datasets to validate the tool call capability of the model trained with \name.\vspace{-0.2cm}
\begin{itemize}
    \item \textbf{BFCL-v2}~\cite{patil2023gorillalargelanguagemodel} primarily consists of Python-style tool call data, divided into four categories \textit{Simple}, \textit{Multiple}, \textit{Parallel}, and \textit{Parallel Multiple}. Version 2 adds more data from dynamic, real world scenarios. We selected the categories that can be evaluated with the Abstract Syntax Tree (AST), which are statistically stable and easy to evaluate.
    The accuracy is reported.\vspace{-0.2cm}
    \item \textbf{API-Bank}~\cite{li-etal-2023-api} is a dialogue-style tool call dataset, including two settings: \textit{Call} and \textit{Retrieve + Call}. The model is required to call predefined local Python tools based on user requirements in the dialogue. Accuracy is measured by evaluating whether the tool return values match the ground truth.\vspace{-0.2cm}
    \item \textbf{ToolAlpaca}~\cite{tang2023toolalpacageneralizedtoollearning} establishes a multi-agent simulator. It utilizes GPT-4~\cite{openai2024gpt4} to simulate the return values of tools. The model can make modifications and re-call the tool based on the return values (e.g., error messages). Finally, GPT-4 evaluates the accuracy of \textit{Process} and \textit{Response}.\vspace{-0.2cm}
\end{itemize}

Additionally, to examine changes in general performance, we evaluated the model's reasoning and conversational abilities using MMLU~\cite{hendrycks2021measuring}, BBH~\cite{suzgun2022challenging}, and MTBench~\cite{10.5555/3666122.3668142}.

\subsubsection{Models}
In our main experiments, we use LLaMA-3.1-8B-Instruct~\cite{dubey2024llama3herdmodels} as base model to examine the effectiveness of the synthetic dialogues generated with \name. For simplicity, we use \name to refer to the fine-tuned model throughout the remainder of this paper.
The models we compared include GPT-3.5, GPT-4, GPT-4o~\cite{openai2024gpt4}, Claude~\cite{bai2022constitutionalaiharmlessnessai}, LLaMA-3.1~\cite{dubey2024llama3herdmodels}, etc, as well as baselines from the paper of the evaluated datasets, such as Lynx-7B~\cite{li-etal-2023-api} and ToolAlpaca-7B~\cite{tang2023toolalpacageneralizedtoollearning}, etc. For specific checkpoint information, please refer to the experimental result tables.

\subsection{Results}
\subsubsection{\name achieves tool-calling capability comparable to GPT-4o.}
We evaluated \name's tool-calling ability on the BFCL. This dataset contains questions from four categories.
In the \textbf{Simple} category, each question contains one tool, which the LLM must correctly call based on requirements. The \textbf{Multiple} question includes 2-4 tools, requiring the model to choose and call the most suitable one. In the \textbf{Parallel} category, several tools should be called in one turn. \textbf{Multiple Parallel} adds distracting candidate tools to the Parallel setup. 

The results are shown in Tabel~\ref{tab:bfcl}. Overall, \name achieves performance comparable to GPT-4o. On the Non-Live subset, \name outperformed GPT-4 and GPT-4o, but was slightly weaker than Claude-3.5-Sonnet. On the Live subset, \name still lags behind these leading closed-source LLMs. 
This is because the Live subset added more user-contributed test cases from the real world, thus making it more challenging for the model.
We attribute this gap primarily to differences in model size, given that \name only has 8B parameters.
% Compared to the baseline, \name achieved performance comparable to most mainstream LLMs. \name slightly outperformed GPT-4 and GPT-4o, but was slightly weaker than Claude-3.5-Sonnet. The difference is mainly reflected in \textit{Parallel Multiple} type questions. 
% In these types of questions, the model is presented with multiple candidate tools and required to call these tools in parallel, which is of high complexity in nature. We believe that the difference in the model sizes results in \name's performance being inferior to Claude, as \name only has 8B parameters.
The ablation experiment shows that the model trained on data synthesized by strategies including both Graph-based Sampling and Generated Plan performs the best. This is especially evident in Parallel Multiple type questions.
% Compared to LLaMA-3.1-8B-Instruct, the trained model achieved an accuracy improvement of 11.5\% on Parallel Multiple (increasing from 73.5\% to 85\%).

While different models or training strategies exhibit some variance in certain categories, the differences in average performance are not significant. Therefore, we conducted further comparative experiments on additional tool call datasets.

\subsubsection{\name achieves SOTA on dialogue data.}

BFCL tests the model's tool calling capability in the form of Q\&A. However, we believe that a conversational format is closer to real-world application scenarios. 
On the other hand, our synthesized training data is also in the form of dialogue. Therefore, in the BFCL test, the advantage of \name cannot be fully demonstrated.

API-Bank is a dialogue dataset. During evaluation, the model needs to make tool call requests after receiving user demands and provides a response based on the tool's return value. This process may occur multiple times within a single dialogue. It includes two test settings: \textit{Call} and \textit{Retrieve + Call}. In the \textit{Call} setting, the assistant selects tools from a candidate tool set to fulfill user requests. In the \textit{Retrieve + Call} setting, the assistant only has access to a search tool. The assistant needs to search for the relevant tools first, and then call them. 

\input{tables/api_bank_noR}

From the results in Table~\ref{tab:api_bank}, \name achieves state-of-the-art average accuracy. Under the \textit{Call} setting, \name outperforms all baselines. In the \textit{retrieve + call} setting, \name is inferior to GPT-3.5-turbo but superior to other baselines, including GPT-4.
The ablation experiments show that Graph-based Sampling strategy can improve the model's accuracy in tool call under this setting. This is because tools obtained through Graph sampling often have sequential correlations. As a result, the training data includes more examples of sequentially calling tools, aligning better with the requirements of the \textit{retrieve + call} setting.

\input{tables/tool_alpaca}

\subsubsection{\name can correct mistakes based on error messages.}
Correcting errors is a key capability of LLM tool calls~\cite{wang2024llmsimaginariumtoollearning}. We conducted the tests in the \textit{Simulated} setting of ToolAlpaca dataset. This dataset established a simulation environment that utilizes GPT-4 to mimic the return values of tools, including the error messages when calls fail. The model is allowed to self-correct based on these error messages and then retry the call. We assess the tool call and correction capabilities of \name on this dataset.

The dataset evaluates the accuracy of the tool call \textit{Procedure} and the model's final \textit{Response}. The procedure is considered accurate when the model's call matches the ground truth. The response is considered accurate when the model's response can satisfy the user’s instruction. If they are both accurate, the model's \textit{Overall} performance is considered accurate. This evaluation was conducted by GPT-4. We presented the results in Table~\ref{tab:tool_alpaca}.

\name's \textit{Procedure} accuracy reached 85\%, surpassing GPT-3.5's 77\%. In \textit{Procedure} evaluation, correcting errors is considered as redundant actions and therefore judged as incorrect. Hence, this accuracy implies that in most cases, the first tool call of \name is accurate. On the other hand, \name's \textit{Response} accuracy of 88\% is higher than the \textit{Procedure} accuracy of 85\%, indicating that \name corrected errors in some test cases. This suggests that \name has the ability to self-correct based on error messages, even though error correction samples are not included in the training data.

\subsubsection{\name's General Ability Is NOT Compromised by Fine-tuning.}
The fine-tuned model risks catastrophic forgetting, where the capability for tool call is enhanced, but other abilities decline. As an AI assistant, LLM's reasoning and conversational abilities are equally important. Therefore, we tested the tuned model on the MMLU, BBH, and MTBench datasets to examine whether catastrophic forgetting issues have occurred. The results are shown in Table~\ref{tab:mtbench_mmlu_bbh}.

The test results on MMLU and BBH show that there is no significant difference in performance between the models before and after training. However, on the MTBench dataset, models trained on data without Graph-based Sampling or Plan-Generation exhibited a decline in performance.
Notably, in the evaluation of \textit{Turn 2}, models trained on synthetic data using the Plan-Generation strategy exhibited a slight performance improvement. This improvement is due to Plan-Generation enhancing the naturalness and coherence of synthetic dialogues, thereby boosting the model's conversational abilities.

\input{tables/mtbench_mmlu_bbh}

%% file: tables/bfcl-v1v2.tex
% Please add the following required packages to your document preamble:
% \usepackage{multirow}
% \usepackage[normalem]{ulem}
% \useunder{\uline}{\ul}{}
\begin{table*}[t]
\setlength\tabcolsep{3pt}
\renewcommand{\arraystretch}{1.1}
\begin{adjustbox}{max width=\textwidth}
\begin{tabular}{cccccccccccccc}
\hline
% \multicolumn{3}{c}{\multirow{2}{*}{\textbf{Baselines}}} 
\multicolumn{3}{c}{}                                             & \multicolumn{4}{c}{\textbf{Non-Live}}                                                     & \multicolumn{4}{c}{\textbf{Live}}                                                    & \multicolumn{3}{c}{\textbf{Overall}}                  \\
% \cdashline{4-14}
\cmidrule(lr){4-7}\cmidrule(lr){8-11}\cmidrule(lr){12-14}
\multicolumn{3}{c}{\textbf{Baselines}}                        & \textit{Simple}      & \textit{Multiple} & \textit{Parallel} & \textit{\begin{tabular}[c]{@{}c@{}}Parallel \\ Multiple\end{tabular}} & \textit{Simple} & \textit{Multiple} & \textit{Parallel} & \textit{\begin{tabular}[c]{@{}c@{}}Parallel \\ Multiple\end{tabular}} & \textit{Non-Live} & \textit{Live}  & \textit{Overall} \\
% \cmidrule(lr){1-3}\cmidrule(lr){4-7}\cmidrule(lr){8-11}\cmidrule(lr){12-14}
\cdashline{1-14}
\multicolumn{3}{l}{\textbf{Claude-3.5-Sonnet}}                   & 88.55                & 95.00             & \textbf{91.50}    & \textbf{92.50}             & 86.82           & 80.06             & 81.25             & 45.83                      & \textbf{91.22}    & 80.75          & 85.98            \\
\multicolumn{3}{l}{\textbf{GPT-4-turbo-0409}}                    & 87.45                & \textbf{96.50}    & 91.00             & 89.00                      & \textbf{87.98}  & \textbf{84.14}    & \textbf{100.00}   & \textbf{79.17}             & 90.28             & \textbf{85.02} & \textbf{87.65}   \\
\multicolumn{3}{l}{\textbf{GPT-4o-0513}}                         & 80.55                & 91.00             & 90.00             & 83.00                      & 81.78           & 77.24             & 87.50             & 75.00                      & 85.02             & 78.20          & 81.61            \\
\multicolumn{3}{l}{\textbf{LLaMA-3.1 8B}}                        & 90.36                & 89.50             & 73.50             & 73.50                      & 74.03           & 73.31             & 56.25             & 54.17                      & 83.40             & 72.88          & 78.14            \\
\rowcolor[HTML]{EFEFEF}                 & \textbf{Graph} & \textbf{Plan}  &                      &                   &                   &                            &                 &                   &                   &                            &                   &                &                  \\
\multirow{4}{*}{\textbf{Ours}} & \textbf{\cmark} & \textbf{\cmark} & 92.18                & 92.50             & {\ul 90.00}       & {\ul 85.00}                & {\ul 73.64}     & {\ul 75.22}       & {\ul 75.00}       & {\ul 70.83}                & {\ul 90.30}       & {\ul 74.83}    & {\ul 82.57}      \\
                               & \textbf{\cmark} & \textbf{\xmark} & 90.73                & {\ul 93.00}       & 89.50             & 84.50                      & 71.71           & 74.35             & {\ul 75.00}       & 66.67                      & 89.60             & 73.71          & 81.66            \\
                               & \textbf{\xmark} & \textbf{\cmark} & 91.45                & {\ul 93.00}       & 88.00             & 84.00                      & 70.93           & 74.73             & 68.75             & 66.67                      & 89.50             & 73.78          & 81.64            \\
                               & \textbf{\xmark} & \textbf{\xmark} & {\ul \textbf{92.36}} & 91.50             & 86.50             & 82.50                      & 72.48           & 73.58             & {\ul 75.00}       & 62.50                      & 89.00             & 73.18          & 81.09            \\ \hline
\end{tabular}
\end{adjustbox}
\caption{Results on BFCL-v2 leaderboard (updated on 08/16/2024). 
"Non-Live" and "Live" indicate the results on v1 and v2 subsets respectively.
Table values are shown as a percentage.
The best results in each category are marked in \textbf{bold}.
The best results from our model are \ul{underlined}.}
\label{tab:bfcl}
\end{table*}

%% file: tables/api_bank_noR.tex
\begin{table}[t]
\centering
\setlength\tabcolsep{4pt}
\begin{tabular}{lccccc}
\hline
\multicolumn{3}{l}{\textbf{Baselines}}                                               & \textbf{L1}            & \textbf{L2}      & \textbf{Avg.}          \\ \hline
\multicolumn{3}{l}{\textbf{GPT-4-turbo-0409}}                                        & 72.43                & 39.26          & 55.85                \\
\multicolumn{3}{l}{\textbf{GPT-4o-0513}}                                             & 76.19                & 42.96          & 59.58                \\
\multicolumn{3}{l}{\textbf{GPT-3.5-turbo-0125}}                                      & 70.43                & \textbf{52.59} & 61.51                \\
\multicolumn{3}{l}{\textbf{Lynx-7B}$^\dagger$}                                                 & 49.87                & 30.37          & 40.12                \\
\multicolumn{3}{l}{{\textbf{Llama-3.1-8B-Instruct}}}                                   & 71.18                & 37.04          & 54.11                \\
\cdashline{1-6}
\rowcolor[HTML]{EFEFEF}\multicolumn{1}{c}{\multirow{5}{*}{\textbf{Ours}}} & \textbf{Graph} & \textbf{Plan}  &                        &                  &                        \\
\cdashline{1-6}
\multicolumn{1}{c}{}                               & \textbf{\cmark} & \textbf{\cmark} & 77.52                 & {\ul 46.68}    & {\ul \textbf{62.10}} \\
\multicolumn{1}{c}{}                               & \textbf{\cmark} & \textbf{\xmark} & 76.26                & 42.23          & 59.25                \\
\multicolumn{1}{c}{}                               & \textbf{\xmark} & \textbf{\cmark} & {\ul \textbf{79.30}} & 38.53          & 58.92                \\
\multicolumn{1}{c}{}                               & \textbf{\xmark} & \textbf{\xmark} & 76.76                & 39.27          & 58.02                \\ \hline
\end{tabular}
\caption{Results on API-Bank Dataset. L1 and L2 refer to the setting of \textit{Call} and \textit{Retrieve + Call}, respectively. Table values are shown as a percentage.
$^\dagger$ indicates results derived from the original paper.}
\label{tab:api_bank}
\vspace{-0.4cm}
\end{table}

%% file: tables/tool_alpaca.tex
\begin{table}[t]
\centering
\setlength\tabcolsep{4pt}
\begin{tabular}{lccccc}
\hline
\multicolumn{3}{l}{\textbf{Baselines}}                                      & \textbf{Proc.} & \textbf{Resp.} & \textbf{Overall} \\ \hline
\multicolumn{3}{l}{\textbf{GPT-3.5}$^\dagger$}                                                 & 77            & 85              & 75              \\
\multicolumn{3}{l}{\textbf{ToolAlpaca-13B}$^\dagger$}                                          & 63            & 69              & 60              \\
\multicolumn{3}{l}{\textbf{ToolAlpaca-7B}$^\dagger$}                                           & 70            & 73              & 70              \\
\multicolumn{3}{l}{\textbf{LLaMA-3.1 8B}}                                                & 74            & 80              & 74              \\ \cdashline{1-6}
\rowcolor[HTML]{EFEFEF}\multicolumn{1}{c}{\multirow{5}{*}{\textbf{Ours}}} & \textbf{Graph} & \textbf{Plan}  &                 &                   &                   \\ \cdashline{1-6}
\multicolumn{1}{c}{}                      & \textbf{\cmark} & \textbf{\cmark} & \textbf{85}   & \textbf{88}     & \textbf{84}     \\
\multicolumn{1}{c}{}                      & \textbf{\cmark} & \textbf{\xmark} & 80            & 85              & 80              \\
\multicolumn{1}{c}{}                      & \textbf{\xmark} & \textbf{\cmark} & 81            & 83              & 79              \\
\multicolumn{1}{c}{}                      & \textbf{\xmark} & \textbf{\xmark} & 78            & 83              & 77              \\ \hline
\end{tabular}
\caption{Results on ToolAlpaca Dataset. \textbf{Proc.} and \textbf{Resp.} stand for \textit{Procedure} and \textit{Response}, respectively. Table values are shown as a percentage.
$^\dagger$ indicates results derived from the original paper.}
\label{tab:tool_alpaca}
\vspace{-0.4cm}
\end{table}

%% file: tables/mtbench_mmlu_bbh.tex
\begin{table}[t]
\centering
\setlength\tabcolsep{1.7pt}
\begin{tabular}{ccccccc}
\hline
\multicolumn{2}{c}{\textbf{Setting}} & \multicolumn{3}{c}{\textbf{MTBench}}              & \multirow{2}{*}{\textbf{MMLU}} & \multirow{2}{*}{\textbf{BBH}} \\ \cmidrule(lr){1-2} \cmidrule(lr){3-5}
\textbf{Graph}    & \textbf{Plan}    & \textit{Turn 1} & \textit{Turn 2} & \textit{Avg.} &                                &                               \\ \hline
\rowcolor[HTML]{EFEFEF}-                 & -                & 7.84            & 6.85            & 7.34          & 69.3\%                          & 63.2\%                          \\
\cmark             & \cmark            & \textbf{8.08}            & \textbf{7.16}            & \textbf{7.62}          & \textbf{69.8\%}                           & 63.3\%                          \\
\cmark             & \xmark            & \textcolor{mypink1}{7.39}            & \textcolor{mypink1}{6.43}            & \textcolor{mypink1}{6.91}          & 69.3\%                           & \textbf{63.5\%}                          \\
\xmark             & \cmark            & \textcolor{mypink1}{7.59}            & 7.04            & \textcolor{mypink1}{7.31}          & \textcolor{mypink1}{69.1\%}                           & 63.2\%                          \\
\xmark             & \xmark            & \textcolor{mypink1}{7.16}            & 6.85            & \textcolor{mypink1}{7.01}          & 68.9\%                           & \textbf{63.5\%}                          \\ \hline
\end{tabular}
\caption{Results of general abilities test on MTBench, MMLU, and BBH. The results in the first row are from the model before tuning.
The best results are marked in \textbf{bold}. 
\textcolor{mypink1}{Red} indicates a decrease in performance.}
\label{tab:mtbench_mmlu_bbh}
\vspace{-0.3cm}
\end{table}

%% file: chapters/analysis.tex
\section{Correlation Analysis}

To further investigate the impact of the diversity and coherence of the dialogue data on model performance, we conducted additional correlation analysis. In the previous experiments, we synthesized a total of $8,000 \times 4 = 32,000$ dialogues by using \name and its ablation settings. We randomly sampled from these data 10 times, each time selecting 4,000 dialogues to form 10 new training sets. We calculated the diversity metrics \textit{D-3} and H, and the coherence metrics \textit{SS} and \textit{EnR} for each dataset. 
Then, we used this data to fine-tune the Llama3.1-8B-Instruct. We tested these ten fine-tuned models on the BFCL and MTBench. 
Finally, we calculated the Pearson correlation coefficient between the evaluation metrics of the training data and the model performance and reported it in Table~\ref{tab:corr}.

The average results on the BFCL show that both diversity and coherence of the training data contribute a lot to enhancing the model's tool-calling capabilities. MTBench results show a strong positive correlation between data coherence and the model's conversational performance, consistent with our assumption. Notably, while we use entropy and Distinct-N scores to assess diversity, the inconsistent correlation between these metrics and model performance suggests they may reflect different dimensions of diversity.
On the other hand, coherence does not appear to positively impact the parallel test sets in the BFCL, likely due to the nature of these tests involving multiple calls within a single turn.
Nevertheless, while our \name has demonstrated the benefits of increasing diversity and coherence in Section~\ref{sec:quality_evaluation}, the correlation results in this section further validate their positive effects on overall performance.

\input{tables/corr}

%% file: tables/corr.tex
\begin{table}[]
\setlength\tabcolsep{2.5pt}
\centering
\begin{tabular}{lllll}
\hline
\textbf{Metrics}           & \multicolumn{1}{c}{\textbf{D-3}} & \multicolumn{1}{c}{\textbf{H}} & \multicolumn{1}{c}{\textbf{SS}} & \multicolumn{1}{c}{\textbf{EnR}} \\
\rowcolor[HTML]{EFEFEF}\multicolumn{5}{c}{\textit{\textbf{BFCL}}}                                                                                                                          \\
Simple            & 0.174                            & \textcolor{mypink1}{ -0.272}  & {\color[HTML]{32CB00} 0.347}    & 0.118                            \\
Parallel          & {\color[HTML]{32CB00} 0.250}     & {\color[HTML]{32CB00} 0.315}   & -0.168                          & 0.072                            \\
Multiple          & {\color[HTML]{32CB00} 0.336}     & -0.042                         & {\color[HTML]{32CB00} 0.644}    & {\color[HTML]{32CB00} 0.612}     \\
Parallel Multiple & \textcolor{mypink1}{ -0.274}    & {\color[HTML]{32CB00} 0.479}   & \textcolor{mypink1}{ -0.310}   & -0.102                           \\
Avg.              & {\color[HTML]{32CB00} 0.250}     & {\color[HTML]{32CB00} 0.273}   & {\color[HTML]{32CB00} 0.255}    & {\color[HTML]{32CB00} 0.378}     \\
\rowcolor[HTML]{EFEFEF}\multicolumn{5}{c}{\textit{\textbf{MTBench}}}                                                                                                                       \\
Turn 1            & -0.087                           & {\color[HTML]{32CB00} 0.221}   & {\color[HTML]{32CB00} 0.262}    & {\color[HTML]{32CB00} 0.298}     \\
Turn 2            & -0.185                           & 0.045                          & {\color[HTML]{32CB00} 0.708}    & {\color[HTML]{32CB00} 0.415}     \\
Avg.              & -0.179                           & 0.146                          & {\color[HTML]{32CB00} 0.650}    & {\color[HTML]{32CB00} 0.454}    \\ \hline
\end{tabular}
\caption{The Pearson correlation between evaluation metrics of data and model's performance. P-values > 0.2 and < 0.2 are marked in {\color[HTML]{32CB00} green} and  \textcolor{mypink1}{red}, respectively.}
\label{tab:corr}
\vspace{-0.4cm}
\end{table}

%% file: chapters/overlap_analysis.tex
\section{Dataset Overlap Analysis}
\label{sec:overlap_analysis}
To ensure the reliability of the evaluation, we conducted an overlap analysis between training and test datasets. This examination helps verify the independence of these test data and prevents potential data leakage issues. We employed both N-gram-based and similarity-based methods to demonstrate that there is no significant data leakage in the \textsc{ToolFlow} dataset. We also included the well-known xLam agent training set~\cite{zhang2024xlamfamilylargeaction} as a control group for comparison.

\textbf{N-gram-based method} Following the approach used in LLaMA-2~\cite{touvron2023llama2openfoundation}, we considered a token contaminated if it appeared in any token n-gram longer than 10 tokens in both the evaluation sample and the training set. A tool was classified as leaked if more than 10\% of the tokens in its JSON string were contaminated.

\textbf{Similarity-based method} We defined a tool as leaked if the cosine similarity between the given tool and any tool in the evaluation dataset exceeded 0.9. We used the 
% \texttt{sentence-transformers/all-MiniLM-L12-v2}
\texttt{all-MiniLM-L12-v2}
encoder from HuggingFace\footnote{\url{https://huggingface.co/}} to obtain representations for all tools.

We present the proportions of data leakage across different evaluation metrics in Table~\ref{tab:dataset_overlap}. These results suggest that there is no severe data leakage between \textsc{ToolFlow} as a training set and the test sets.
\input{tables/overlap_table}

%% file: tables/overlap_table.tex
\begin{table}[h]
\centering
\setlength\tabcolsep{2.4pt}
\begin{adjustbox}{max width=\columnwidth}
\begin{tabular}{lcccc}
\hline
\multicolumn{1}{r}{\textbf{Training}} & \multicolumn{2}{c}{\textbf{ToolFlow}}                                                                           & \multicolumn{2}{c}{\textbf{xLam}}                                                                               \\ \cmidrule(lr){2-3}\cmidrule(lr){4-5}
\textbf{Test}                         & \textit{N-gram}                                        & \textit{Similarity}                                    & \textit{N-gram}                                        & \textit{Similarity}                                    \\ \hline
BFCL                                  & \cellcolor[HTML]{FFFDFA}{\color[HTML]{333333} 0.239\%} & \cellcolor[HTML]{FFFDFA}{\color[HTML]{333333} 2.922\%} & \cellcolor[HTML]{FFFDFA}{\color[HTML]{333333} 0.656\%} & \cellcolor[HTML]{FFFDFA}{\color[HTML]{333333} 5.247\%} \\
APIBank                               & 0.000\%                                                & \cellcolor[HTML]{FFFDFA}{\color[HTML]{333333} 1.887\%} & 0.000\%                                                & \cellcolor[HTML]{FFFDFA}{\color[HTML]{333333} 3.774\%} \\
ToolAlpaca                            & \cellcolor[HTML]{FFFDFA}{\color[HTML]{333333} 0.000\%} & 0.000\%                                                & 0.000\%                                                & 0.000\%                                                \\ \hline
\end{tabular}
\end{adjustbox}
\caption{Overlap between Training and Test Sets}
\label{tab:dataset_overlap}
% \vspace{}
\end{table}

%% file: chapters/conclusion.tex
\section{Conclusion}
In this work, we propose Graph-based Sampling and Planned Generation strategies to enhance the diversity and coherence of synthetic data. Based on these two strategies, we introduce a pipeline called \name for synthesizing tool calling data and generate 8,000 training samples. Using this dataset, we conduct SFT on Llama3.1-8B-Instruct, resulting in improved tool calling capability of the model. Subsequently, we conduct correlation analysis to demonstrate the influence of data diversity and coherence on model performance. This provides a reference for the composition of training data for the tool-enhanced agent.

%% file: chapters/limitation.tex
\section*{Limitations}
We summarize the limitations in two points. 

As described in Section~\ref{sec:imp_details}, the seed data is a pre-collected tool set including 16,000 APIs. Although our \name can synthesize more diverse data, it is undeniable that the size and diversity of the tool set also affect the diversity of the data. However, how to enrich the seed data has not yet been studied in this work.

On the other hand, \name utilizes GPT-4 for data synthesis, and then uses this data to train a 8B-model. Therefore, it still falls under the paradigm of using strong models to train weak models. Whether the model can be improved by training on its own synthesized data is still unknown. We believe that this weak-to-strong setting is more challenging but also more meaningful.

%% file: chapters/ethic_statement.tex
\section*{Ethic Statement}
In this research, GPT-4 was employed as an evaluator and generator in a manner consistent with ethical guidelines. Transparency about its usage, accountability for its outputs, and mitigation of potential biases were prioritized. Data privacy and security were strictly maintained, and the AI's limitations were acknowledged, ensuring it supplemented rather than replaced human judgment. This approach aimed to enhance the research quality while upholding academic integrity and ethical standards. 

%% file: chapters/acknowledgement.tex
\section*{Acknowledgements}
This work was partially supported by Hong Kong RGC GRF No. 14206324, CUHK direct grant No. 4055209, and CUHK Knowledge Transfer Project Fund No. KPF23GWP20.

%% file: chapters/appendix.tex
\clearpage
\appendix

\section{Appendix}
\label{sec:appendix}
\subsection{Details of Data Quality Assessment}
\label{sec:data_quality_assess_app}
Following~\citet{dziri-etal-2019-evaluating}, we converted the dialogue data into a Natural Language Inference (NLI) format. In this format, the request and response from the previous dialogue round serve as the \textbf{premise}, and the current round's request serves as the \textbf{hypothesis}. We then use a trained BERT~\cite{reimers2019sentencebertsentenceembeddingsusing} to predict the relationship between the two, and we calculate the ratio of entailment predictions (EnR). A higher proportion indicates greater coherence between consecutive dialogue rounds. Additionally, we measure the semantic similarity (SS) between the premise and hypothesis. We extract sentence representations using BERT and compute their cosine similarity. A higher similarity score indicates a more coherent dialogue.

Regarding diversity, we calculate the text's Shannon entropy (\textit{H}) based on the word frequency. 
We also compute the Distinct-N Score~\cite{li-etal-2016-diversity} for the dataset, with $N=3$ (\textit{D-3}). Higher entropy or Distinct-N Score indicates that the dataset contains more information and has greater diversity. In addition, we sampled 200 dialogues in each set. We used GPT-4 to carefully evaluate each dialogue based on four dimensions: naturalness (\textit{NAT}), coherence (\textit{COH}), helpfulness (\textit{HELP}), and accuracy (\textit{ACC}). The prompt for GPT-4 evaluation is shown in Table~\ref{tab:eval_prompt}.

Automatic evaluation indicates that the Planned-Generation strategy enhances conversation coherence. Both coherence metrics, SS and EnR, reflect this improvement. Intuitively, the plan is carefully designed by the model in advance, leading to more coherent dialogue. On the other hand, the Graph Sampling strategy can increase data diversity. This is because the strategy samples tools with strong associations, and the combination of these tools enhances data diversity.

GPT-4's evaluation indicates that the Planned-Generation strategy enhances the naturalness of dialogue. This metric assesses whether a dialogue could realistically occur in the real world. In data synthesized without a plan, most user requests are tool calls with little chitchat, which is uncommon in real-world scenarios, resulting in lower naturalness. GPT-4's coherence evaluation closely aligns with automatic assessments. In terms of helpfulness, both Graph Sampling and Planned-Generation strategies show improvement. The low scores in Helpfulness are mainly due to the assistant frequently asking follow-up questions about parameters, which consumes dialogue turns. These strategies help reduce such behavior. The accuracy of synthesized data is high across all four settings, likely due to pre-filtering with quality control tools.

\subsection{Reliability Analysis of GPT-4 as an Evaluator}
\label{sec:gpt_4_evaluate}
% \subsubsection{Human Evaluation}
To validate the reliability of GPT-4's evaluations, we conducted additional human evaluations. Specifically, we sampled 50 examples from the data previously evaluated by GPT-4 for human assessment. We recruited four Computer Science PhD students to rate these 50 samples using the same criteria as GPT-4: naturalness, coherence, helpfulness, and accuracy. We collected the human evaluation results and calculated both the Cohen's Kappa inter-rater agreement scores among evaluators and the Pearson correlation coefficients between human and GPT-4 evaluations.
\input{tables/human_eval}

The results in Table~\ref{tab:human_eval} demonstrate high inter-rater agreement and strong correlations with GPT-4's evaluations, particularly for helpfulness and accuracy metrics. While the agreement and correlations for naturalness and coherence were relatively lower compared to the other two metrics, they still maintained a minimum correlation of 0.61, indicating a strong positive correlation between human and GPT-4 evaluations. We believe these results substantiate the reliability of GPT-4's evaluations.

\subsection{GPT-4 v.s. Open Source LLM for Data Synthesis}
To evaluate the reliance of synthesis on closed-source, high-performance LLMs (e.g., GPT-4), we conducted comparative experiments using open-source alternatives. Following the same pipeline but replacing GPT-4 with LLaMA-3.1-8B-instruct, we conducted preliminary synthesis experiments. While still in the initial validation phase, we have synthesized 4,235 dialogue instances. We replicated experiments using this data to fine-tune LLaMA-3.1-8B-instruct, with BFCL test results shown in Table~\ref{tab:comp_gpt4_llama}.
\input{tables/llama31_syn}
\input{tables/other_model_bfcl}
\input{tables/other_model_api_bank}
Here, "N/A" represents the baseline performance of LLaMA-3.1-8B-instruct (consistent with results reported in the main text). Notably, we leveraged existing results from Section 6's Correlation analysis, where we had downsampled to ten subsets of 4,000 dialogues each, comparable to our LLaMA-3.1-8B-instruct synthesized dataset size. The table reports the average results across these ten subsets. While data synthesized by LLaMA-3.1-8B-instruct indeed shows lower performance compared to GPT-4-synthesized data, the gap is not substantial. Moreover, compared to LLaMA-3.1-8B-instruct's initial performance, training on its self-synthesized data demonstrates improvements in tool-calling capabilities.

\subsection{Train with other base model with \name}
Table \ref{tab:other_model_bfcl} and \ref{tab:other_model_api_bank} display the results of two other base models fine-tuned with data generated with \name. Results demonstrate that \textsc{ToolFlow} enhances tool-calling performance across multiple LLMs, validating the generalizability of our synthetic data approach.

\subsection{Prompts and Demonstrations}
See Table~\ref{tab:plan_prompt} to~\ref{tab:tool_prompt} for details.

\begin{table*}[]
    \centering
    % left, bottom, right, upper
    \includegraphics[trim={3cm 15.2cm 3cm 0cm}, clip, width=1.0\textwidth]{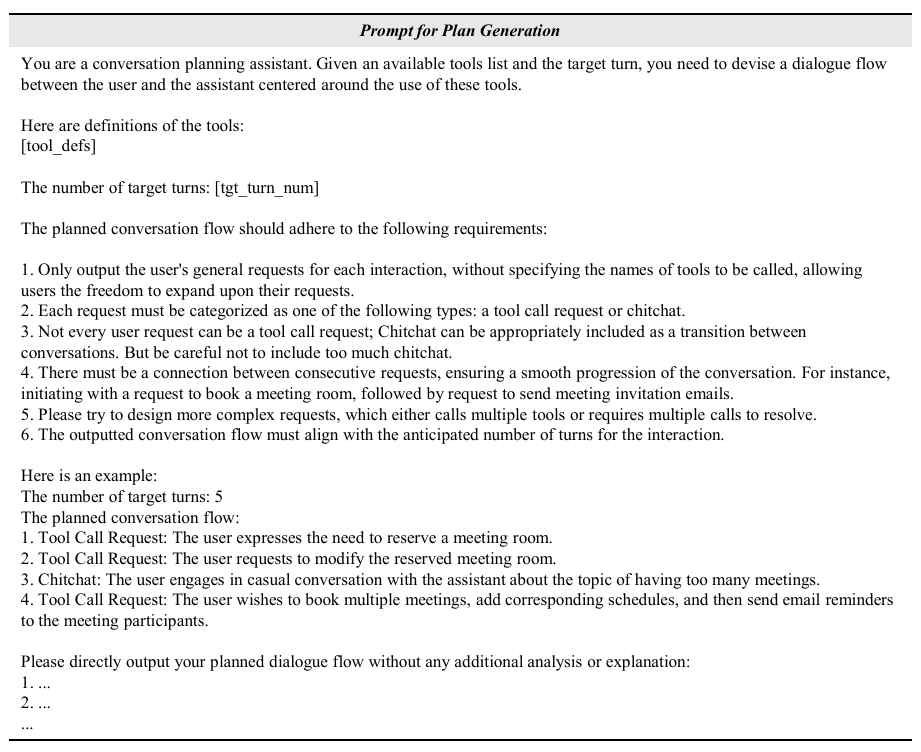}
    \caption{Prompt for plan generation}
    \label{tab:plan_prompt}
    % \vspace{-2mm}
\end{table*}

\begin{table*}[]
    \centering
    % left, bottom, right, upper
    \includegraphics[trim={3cm 15.5cm 3cm 0cm}, clip, width=1.0\textwidth]{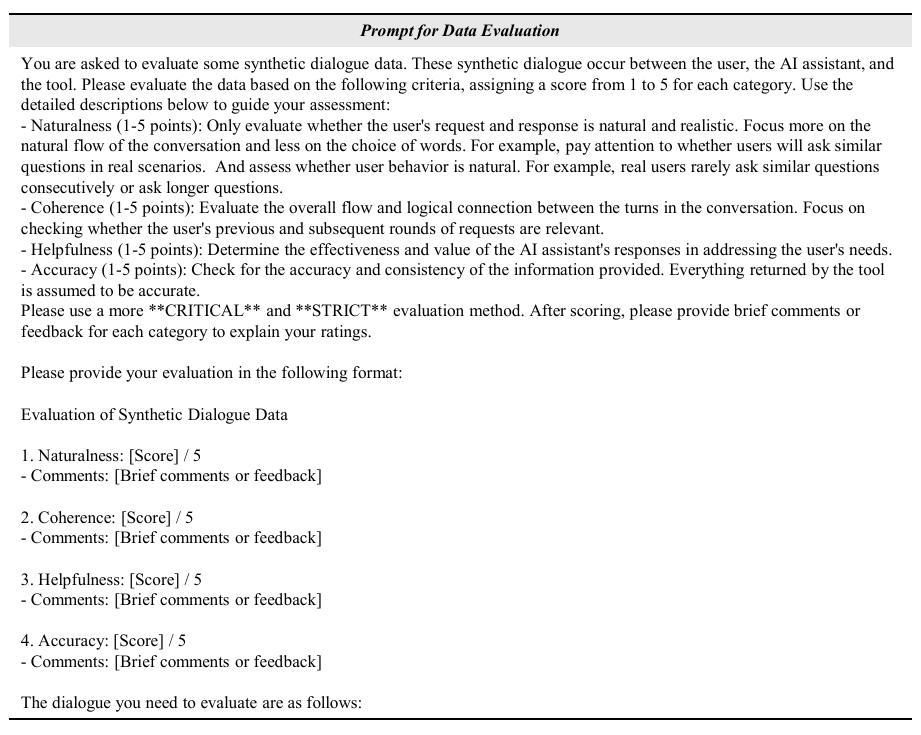}
    \caption{Prompt for GPT-4 Data Evaluation}
    \label{tab:eval_prompt}
    % \vspace{-2mm}
\end{table*}

\begin{table*}[]
    \centering
    % left, bottom, right, upper
    \includegraphics[trim={3cm 5.5cm 3cm 0cm}, clip, width=1.0\textwidth]{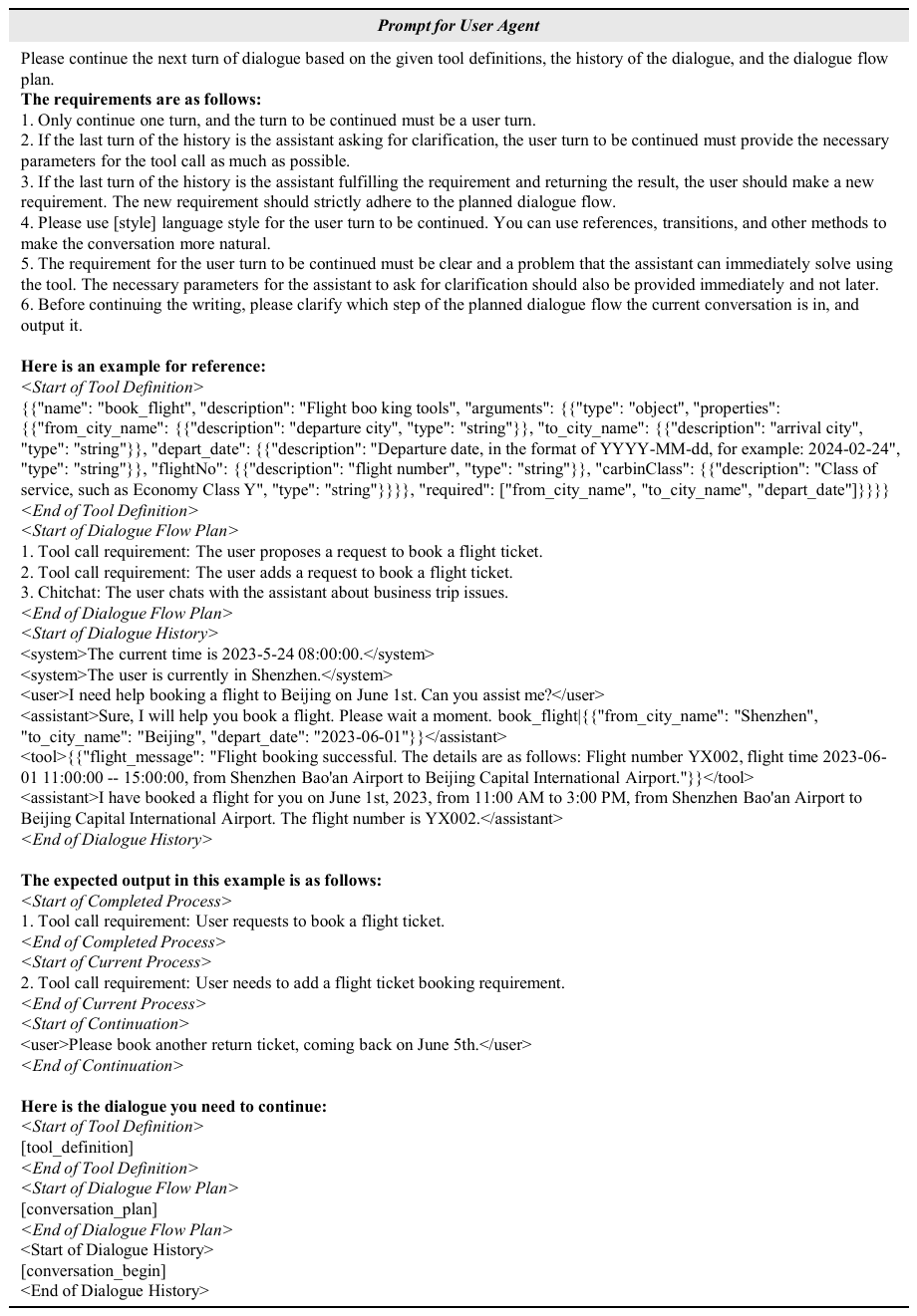}
    \caption{Prompt for User agent}
    \label{tab:user_prompt}
    % \vspace{-2mm}
\end{table*}

\begin{table*}[]
    \centering
    % left, bottom, right, upper
    \includegraphics[trim={3cm 16.5cm 3cm 0cm}, clip, width=1.0\textwidth]{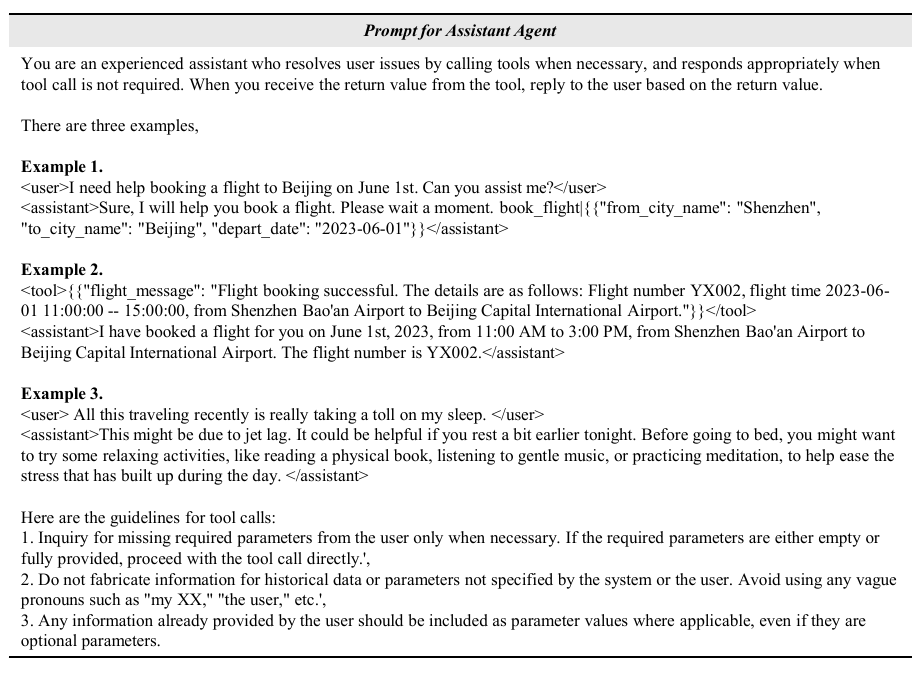}
    \caption{Prompt for Assistant agent}
    \label{tab:assistant_prompt}
    % \vspace{-2mm}
\end{table*}

\begin{table*}[]
    \centering
    % left, bottom, right, upper
    \includegraphics[trim={3cm 18.7cm 3cm 0cm}, clip, width=1.0\textwidth]{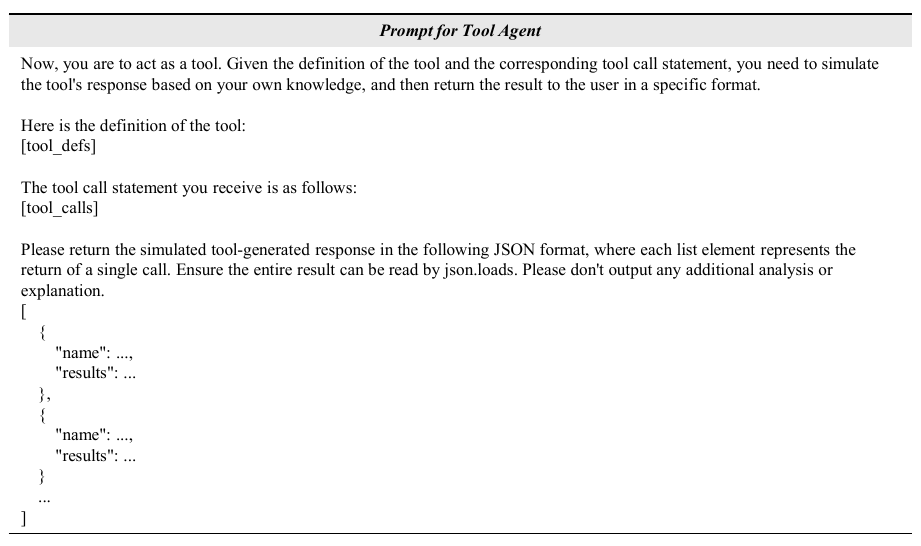}
    \caption{Prompt for Tool agent}
    \label{tab:tool_prompt}
    % \vspace{-2mm}
\end{table*}

\begin{figure*}[t]
\centering
\begin{minipage}{0.9\textwidth}
\centering
\begin{lstlisting}[language=json]
{
    "type": "function",
    "function": {
        "name": "getcurrency",
        "description": "Get the current exchange rate for a specific currency pair",
        "parameters": {
            "type": "object",
            "properties": {
                "basecurrency": {
                    "type": "string",
                    "description": "The base currency code, e.g., USD"
                },
                "targetcurrency": {
                    "type": "string",
                    "description": "The target currency code, e.g., EUR"
                }
            },
            "required": ["basecurrency", "targetcurrency"]
        },
        "results": {
            "type": "object",
            "properties": {
                "exchangerate": {
                    "type": "number",
                    "description": "The current exchange rate from base currency to target currency"
                },
                "last_updated": {
                    "type": "string",
                    "description": "The date and time when the exchange rate was last updated"
                }
            }
        }
    }
}   
\end{lstlisting}
\end{minipage}
\caption{Example tool in JSON format.}
\label{fig:tool_demo}
\end{figure*}

%% file: tables/human_eval.tex
\begin{table}[h]
\centering
\setlength\tabcolsep{3pt}
\begin{tabular}{lllll}
\hline
\textbf{Metric}     & \textbf{NAT} & \textbf{COH} & \textbf{HELP} & \textbf{ACC} \\ \hline
Cohen's Kappa       & 0.63         & 0.71         & 0.92          & 0.95         \\
Pearson Correlation & 0.61         & 0.76         & 0.84          & 0.86         \\ \hline
\end{tabular}
\caption{Correlation between Human and GPT-4 Ratings}
\label{tab:human_eval}
\end{table}

%% file: tables/llama31_syn.tex
\begin{table*}[]
\centering
\begin{tabular}{llllll}
\hline
\textbf{Data Source}  & \textbf{Simple} & \textbf{Multiple} & \textbf{Parallel} & \textbf{Parallel multiple} & \textbf{Avg.} \\ \hline
N/A                   & 90.36           & 89.50             & 73.50             & 73.50                      & 83.40         \\
LLaMA-3.1-8B-instruct & 89.88           & 90.10             & 85.55             & 80.15                      & 86.42         \\
GPT-4                 & 91.23           & 91.85             & 87.10             & 84.45                      & 88.66         \\ \hline
\end{tabular}
\caption{Performance comparison of models trained on data synthesized by GPT-4 and LLaMA.}
\label{tab:comp_gpt4_llama}
\end{table*}

%% file: tables/other_model_bfcl.tex
\begin{table*}[t]
\centering
\begin{tabular}{lccccc}
\hline
\multicolumn{1}{c}{\textbf{Models}}   & \textbf{Simple} & \textbf{Multiple} & \textbf{Parallel} & \textbf{Parallel Multiple} & \textbf{Avg.} \\ \hline
\textbf{Mistral-7B-Instruct-v0.1}     & 61.27\%         & 54.00\%           & 50.50\%           & 47.50\%                    & 53.32\%       \\
\multicolumn{1}{r}{\textbf{\textsc{+ ToolFlow}}} & 86.91\%         & 81.00\%           & 89.00\%           & 80.50\%                    & 84.35\%       \\
\textbf{Qwen2-7B}                     & 76.73\%         & 63.50\%           & 82.00\%           & 55.50\%                    & 69.43\%       \\
\multicolumn{1}{r}{\textbf{\textsc{+ ToolFlow}}} & 90.18\%         & 81.50\%           & 89.00\%           & 78.50\%                    & 84.80\%       \\ \hline
\end{tabular}
\caption{The results of Mistral 7B and Qwen2 7B on BFCL. '\textsc{+ToolFlow}' represents the results of the model after training with our data.}
\label{tab:other_model_bfcl}
\end{table*}

%% file: tables/other_model_api_bank.tex
\begin{table}[]
\centering
\setlength\tabcolsep{3pt}
\begin{tabular}{lccc}
\hline
\multicolumn{1}{c}{\textbf{Models}}     & \textbf{L1} & \textbf{L2} & \textbf{Avg.} \\ \hline
\textbf{Mistral-7B-Instruct-v0.1}       & 59.14     & 32.48     & 45.81       \\
\multicolumn{1}{r}{\textbf{+ \textsc{ToolFlow}}} & 68.42     & 45.19     & 56.81       \\
\textbf{Qwen2-7B}                       & 58.63     & 25.19     & 41.91       \\
\multicolumn{1}{r}{\textbf{+ \textsc{ToolFlow}}} & 64.97     & 36.93     & 50.95       \\ \hline
\end{tabular}
\caption{The results of Mistral 7B and Qwen2 7B on the API-Bank dataset. '\textsc{+ToolFlow}' represents the results of the model after training with our data.}
\label{tab:other_model_api_bank}
\end{table}